\documentclass[10pt,conference,a4paper]{IEEEtran}

\usepackage{float}
\usepackage{times}
\usepackage{epsfig}
\usepackage{graphicx}
\usepackage{amsmath}
\usepackage{amssymb}
\usepackage{algorithmic}
\usepackage{algorithm}
\usepackage{color}
\usepackage{subfigure}
\usepackage{mathrsfs}
\usepackage{bm}
\usepackage{url}
\usepackage{array}
\usepackage{makecell}

\definecolor{fgreen}{RGB}{34,139,34}

\def\eg{\emph{e.g.}}
\def\ie{\emph{i.e.}}

\def\etc{\emph{etc.}}

\begin{document}

\title{Multiple Instance Dictionary Learning using Functions of Multiple Instances}
\author{\IEEEauthorblockN{Changzhe Jiao}
	\IEEEauthorblockA{ Electrical \& Computer Engineering\\
		University of Missouri\\
		Email: cjr25@mail.missouri.edu}
	\and
	\IEEEauthorblockN{Alina Zare}
	\IEEEauthorblockA{ Electrical \& Computer Engineering\\
		University of Missouri\\
		Email: zarea@missouri.edu}}

\maketitle
%
\begin{abstract}
 Dictionary Learning Functions of Multiple Instances (DL-FUMI)  is proposed to address target detection problems with inaccurate training labels.   DL-FUMI is a multiple instance dictionary learning method that estimates target atoms that describe distinctive and representative features of the target class and background atoms that account for the shared features found across both target and non-target data points. Experimental results show that the target atoms estimated by DL-FUMI are more discriminative and representative of the target class than comparison methods.  DL-FUMI is shown to have improved performance on several detection problems as compared to other multiple instance dictionary learning algorithms.
\end{abstract}

\section{Introduction}
\label{sec:intro}
Obtaining accurate training label information is often time consuming, expensive, and/or infeasible for large data sets.   Furthermore, annotators may be inconsistent during labeling providing inherently imprecise labels. Thus, in many applications, one has access only to inaccurately or weakly labeled training data.     

Sparse coding and dictionary learning methods, whose low-rank data representations generally reduce redundancy and improve discrimination ability, have been successfully applied to many applications \cite{mairal2014sparsemodeling}.  
DL-FUMI leverages the benefits of discriminative dictionary learning for  target detection applications given only inaccurate labels.  This is accomplished through the use of a novel model that assumes each target data point is a \emph{mixture} of both target and background atoms whereas non-target data points are composed of only background atoms.   In other words, unlike the majority of discriminative dictionary learning methods, DL-FUMI does \underline{not} learn a separate dictionary for each class.  Instead, DL-FUMI introduces a shared background dictionary that is used in reconstruction of both target and non-target points. The advantage of this model over class-specific dictionaries is that the target atoms only need to account for the unique characteristics of target (and do not need to address any shared background variability) resulting in more discriminative, representative target atoms.


 Furthermore, the target atoms estimated by DL-FUMI can be examined to uncover what discriminates target data points. Since most approaches estimate class-specific dictionaries, each dictionary must characterize both the class-specific characteristics and the characteristics shared among all data points.  Thus, in these methods, it is often difficult to pin down what is unique about each class without prior insight since the class-specific features are mixed with background features and spread across the atoms.  In contrast, DL-FUMI provides that insight by pulling out the unique target characteristics and identifying which atoms contain those characteristics.  In summary, DL-FUMI advances discriminative dictionary learning by (1) addressing multiple instance learning problems and (2) using a shared background model resulting in improved target characterization and discrimination.



\subsection{Multiple-instance learning (MIL): } MIL \cite{Dietterich:1997} is a variation on supervised learning for problems with inaccurate label information. In particular, training data is segmented into positive and negative \textit{bags}. A bag is defined to be a multi-set of data points. In the case of target detection, the MIL problem requires that a positive bag contains at least one instance from the target class and negative bags are composed of entirely non-target data. Given training data of this form, the overall goal can be to predict either unknown instance-level or bag-level labels on test data. MIL methods are effective for problems where accurately  labeled training data is unavailable.

Most MIL approaches focus on learning a classification decision boundary to distinguish between positive and negative instances/bags \cite{andrews2002support,chen2006miles}. 
%
%
 Although these decision boundary approaches are effective at training classifiers given inaccurate labels, they do not provide an intuitive description or \textit{representative concept} that characterizes the salient and discriminative features of the target class.  The approaches that estimate target representatives \cite {Maron:1998, Zhang:2002, Zare:2015fumi} often only find a single target concept and are, thus, unable to account for large variation in the target class. To address this, DL-FUMI learns a set of target atoms (and background atoms) to characterize target variation. 
\begin{figure*}[!t]
	\vspace{+3mm}
	\begin{equation}
	F=\frac{1}{2}\sum_{i=1}^Nw_i\bigg\|(\mathbf{x}_i-z_i\sum_{t=1}^T\alpha_{it}\mathbf{d}_t^+-\sum_{k=1}^M\alpha_{ik}\mathbf{d}_k^-)\bigg\|_2^2+\lambda\sum_{i=1}^{N}w_i\left\|\begin{bmatrix}z_i \boldsymbol{\alpha}^+_i \\ \boldsymbol{\alpha}^-_i \end{bmatrix} \right\|_1+\sum_{k=1}^M\sum_{t=1}^{T}\gamma_{kt}\langle\mathbf{d}_k^-,\mathbf{d}_{t^{\text{old}}}^+\rangle
	\label{eqn:gFUMI}\tag{3}
	\end{equation}

	\begin{small}
		\begin{equation}
		E[F]=\sum_{\substack{z_i\in\{0,1\}}} P(z_i|\mathbf{x}_i, \boldsymbol{\theta}^{(l-1)})  \left[ \frac{1}{2}\sum_{i=1}^N w_i \left\| \mathbf{x}_i - z_i\sum_{t=1}^T\alpha_{it}\mathbf{d}_t^+ - \sum_{k=1}^M\alpha_{ik}\mathbf{d}_k^-\right\|_2^2+\lambda\sum_{i=1}^{N}w_i\left\|\begin{bmatrix} z_i \boldsymbol{\alpha}^+_i \\ \boldsymbol{\alpha}^-_i \end{bmatrix} \right\|_1\right]+\sum_{k=1}^M\sum_{t=1}^{T}\gamma_{kt}\langle\mathbf{d}_k^-,\mathbf{d}_{t^{\text{old}}}^+\rangle
		\label{eqn:E_gFUMI}\tag{4}
		\end{equation}
		\hrulefill
	\end{small}
\end{figure*}

%

\subsection{Supervised Dictionary Learning: } Sparse coding refers to the task of decomposing a signal into a sparse linear combination of dictionary atoms \cite{mallat1993matching, donoho2006compressed}.  
%
Of particular relevance are supervised (i.e., task-driven or discriminative) dictionary learning methods \cite{mairal2012task,jiang2013label}.  However, among supervised dictionary learning methods, there are only a few approaches that address the problem given inaccurate MIL labels.
 These include MMDL \cite{wang2013max} that trains many linear SVM classifiers and views the estimated parameters as dictionary atoms and  DMIL \cite{shrivastava2014dictionary,shrivastava2015gen} that learns class-specific dictionaries by maximizing the noisy-OR model in such a way that the all negative instances are poorly represented by the estimated target dictionary.
 As outlined in Sec \ref{sec:intro}, DL-FUMI is unique from these existing methods through the use of a shared background dictionary.   


\section{DL-FUMI} \label{sec:3_gFUMI}
Let $\mathbf{X}=\left[\mathbf{x}_1,\cdots,\mathbf{x}_N\right]\in\mathbb{R}^{d\times N}$ be training data where $d$ is the dimensionality of an instance, $\mathbf{x}_i$, and $N$ is the total number of training instances. The data is grouped into $K$ \textit{bags},  $\mathbf{B} = \left\{ \mathbf{B}_1, \ldots, \mathbf{B}_K\right\}$, with associated binary bag-level labels, $L = \left\{L_1, \ldots, L_K\right\}$ where $L_j \in \left\{ 0, 1\right\}$ and $\mathbf{x}_{ji} \in \mathbf{B}_j$ denotes the $i^{th}$ instance in bag $\mathbf{B}_j$.
Given training data in this form, DL-FUMI models each instance as a sparse linear combination of target and/or background atoms $\mathbf{D}$, $\mathbf{x}_i\approx\mathbf{D}\boldsymbol{\alpha}_i$, where $\boldsymbol{\alpha}_i$ is the sparse vector of  weights for instance $i$. Positive bags (\ie, $\mathbf{B}_j$ with $L_j = 1$, denoted as $\mathbf{B}_j^+$) contain at least one instance composed of some target:

\vspace{-3mm}
\begin{eqnarray}
&&\text{if }L_j = 1,  \nonumber \exists \mathbf{x}_i \in \mathbf{B}_j^+ \text{ s.t. } \\
&&\mathbf{x}_i = \sum_{t=1}^{T}\alpha_{it}\mathbf{d}_t^+ + \sum_{k=1}^{M} \alpha_{ik}\mathbf{d}_{k}^-+\boldsymbol{\varepsilon}_{i}, \alpha_{it} \ne 0,
\label{eq:l1}
\end{eqnarray}
where $\boldsymbol{\varepsilon}_i$ is a noise term.   However, the number of instances in a positive bag with a target component is unknown.

If $\mathbf{B}_j$ is a negative bag (\ie, $L_j = 0$, denoted as $\mathbf{B}_j^-$), then this indicates that $\mathbf{B}_j^-$ does not contain any target:
\vspace{-2mm}
\begin{equation}
\text{if }L_j = 0,  \forall \mathbf{x}_i \in \mathbf{B}_j^-, \mathbf{x}_i =  \sum_{k=1}^{M} \alpha_{ik}\mathbf{d}_{k}^-+\boldsymbol{\varepsilon}_{i}
 \label{eq:l2}
\end{equation}
\vspace{-2mm}

Given this problem formulation, the goal of DL-FUMI is to estimate the dictionary\footnotemark[1] $\mathbf{D}=\begin{bmatrix}\mathbf{D}^+ & \mathbf{D}^-\end{bmatrix}\in\mathbb{R}^{d\times (T+M)}$, where $\mathbf{D}^+ = \left[\mathbf{d}_{1}^+,\cdots,\mathbf{d}_{T}^+\right]$ are the $T$ target  atoms and $\mathbf{D}^- = \left[\mathbf{d}_{1}^-,\cdots,\mathbf{d}_{M}^-\right]$ are the $M$ background  atoms. This is accomplished by minimizing \eqref{eqn:gFUMI} which is proportional to the {complete} negative data log-likelihood, where $\boldsymbol{\alpha}_{i}^{l+}$ and $\boldsymbol{\alpha}_{i}^{l-}$ are subsets of $\boldsymbol{\alpha}_{i}$ corresponding to $\mathbf{D}^+$ and $\mathbf{D}^-$, respectively. 
The first term in \eqref{eqn:gFUMI} computes the squared residual error between each instance and its estimate using the dictionary.  In this term, a set of hidden binary latent variables $\left\{z_i\right\}_{i=1}^{N}$ that indicate whether an instance is or is not a target (\ie, $z_i = 1$ when $\mathbf{x}_i$ contains target) are introduced.  For all points in negative bags, $z_i = 0$.  For points in positive bags, the value of $z_i$ is unknown.   Also, a weight $w_i$ is included where $w_i= 1$ if $\mathbf{x}_i \in \mathbf{B}_j^-$ and $w_i = \psi$ if $\mathbf{x}_i \in \mathbf{B}_j^+$ where $\psi$ is a fixed parameter.  This weight helps balance terms when there is a large imbalance between the number of negative and positive instances. 
\footnotetext[1]{$\begin{bmatrix} \mathbf{A} & \mathbf{B}\end{bmatrix}$ and $\begin{bmatrix} \mathbf{A} \\ \mathbf{B}\end{bmatrix}$ are the concatenation of arrays $\mathbf{A}$ and $\mathbf{B}$ horizontally and vertically, respectively.}

%
%

The second term is an $l_1$ regularization term to promote sparse weights. It also includes the latent variables, $z_i$, to account for the uncertain presence of target in positive bags.
	
The third term is a {robust penalty term} that promotes discriminative target atoms (and inspired by a term presented in \cite{ramirez2010classification}). Instead of using a fixed penalty coefficient, we introduce an adaptive coefficient $\gamma_{kt}$ defined as:
\setcounter{equation}{4}
\begin{equation}
\gamma_{kt}=\Gamma\frac{\langle\mathbf{d}_{k}^-,\mathbf{d}_{t}^+\rangle}{\|\mathbf{d}_{k}^-\|\|\mathbf{d}_{t}^+\|}=\Gamma\cos\theta_{kt},
\label{eqn:robust_t} 
\end{equation}%
where $\theta_{kt}$ is the vector angle between the $k^{th}$ background atom and the $t^{th}$ target  atom. Since $sign(\gamma_{kt}) = sign(\langle\mathbf{d}_{k}^-,\mathbf{d}_{t}^+\rangle)$, this discriminative term is always positive and will add large penalty when $\mathbf{d}_k^-$ and $\mathbf{d}_t^+$ have similar shape.  Thus, this term encourages a discriminative dictionary by promoting background atoms that are orthogonal to  target atoms. In implementation, $\gamma_{kt}$ is updated once per iteration using $\mathbf{d}_{k^{old}}^-$ and $\mathbf{d}_{t^{old}}^+$ which are the dictionary values from the previous iteration. 
\vspace{-2mm}
\section{DL-FUMI Optimization} \label{sec:4_optimization}
Expectation-Maximization is used to optimize \eqref{eqn:gFUMI} and estimate $\mathbf{D}$.  During optimization, the fact that many of the binary latent variables $\left\{z_i\right\}_{i=1}^N$ are unknown is addressed by taking the expected value of the log likelihood with respect to $z_i$ as shown in \eqref{eqn:E_gFUMI}.  In  \eqref{eqn:E_gFUMI}, $\boldsymbol{\theta}^{l} = \left\{ \mathbf{D}, \left\{ \boldsymbol{\alpha}_i \right\}_{i=1}^N \right\}$is the set of parameters estimated at iteration $l$ and  $P(z_i|\mathbf{x}_i, \boldsymbol{\theta}^{(l-1)})$ is the probability that each instance is or is not a true target instance. During the E-step of each iteration, $P(z_i|\boldsymbol{x}_i, \boldsymbol{\theta}^{(l-1)})$ is computed as: 
\begin{eqnarray}
& &P(z_i|\mathbf{x}_i, \boldsymbol{\theta}^{(l-1)}) = \nonumber  \\
& &\left\{ \begin{array}{l l}
e^{-\beta \left\| \mathbf{x}_i - \sum_{k=1}^M\alpha_{ik}\mathbf{d}_k^-\right\|_2^2} & \text{if } z_i = 0, L_j = 1\\
1-e^{-\beta \left\| \mathbf{x}_i - \sum_{k=1}^M\alpha_{ik}\mathbf{d}_k^-\right\|_2^2} & \text{if } z_i = 1, L_j = 1\\
0 &  \text{if } z_i = 1, L_j = 0\\
1 &  \text{if } z_i = 0, L_j = 0\\
\end{array}\right.,
\label{eqn:pz1} 
\end{eqnarray}    where $\beta$ is a fixed scaling parameter.
If $\mathbf{x}_i$ is a non-target instance, then it should be characterized by the background atoms well, thus $P(z_i=0|\mathbf{x}_i, \boldsymbol{\theta}^{(l-1)}) \approx 1$. Otherwise, if $\mathbf{x}_i$ is a true target instance, it will not be characterized well using only the background atoms and $P(z_i=1|\mathbf{x}_i, \boldsymbol{\theta}^{(l-1)}) \approx 1$. 

\vspace{.15cm}
\vspace*{+.15cm}
\begin{algorithm}
\caption{DL-FUMI EM algorithm}
\algsetup{indent=2em}
\begin{algorithmic}[1] 
\STATE Initialize $\boldsymbol{\theta}^{0} = \left\{ \mathbf{D}, \left\{ \boldsymbol{\alpha}_i \right\}_{i=1}^N  \right\}$, $l = 1$
\REPEAT
\STATE \textbf{\emph{E-step}}: Compute  $P(z_i|\mathbf{x}_i, \boldsymbol{\theta}^{(l-1)})$
\STATE \textbf{\emph{M-step}}:
        \STATE Update $\mathbf{d}_t^+$ using \eqref{eqn:update_dt}, $\mathbf{d}_t^+\gets\frac{1}{\|\mathbf{d}_t^+\|_2}\mathbf{d}_t^+, t=1,\cdots,T$
        \STATE Update $\mathbf{d}_k^-$ using \eqref{eqn:update_dk}, $\mathbf{d}_k^-\gets\frac{1}{\|\mathbf{d}_k^-\|_2}\mathbf{d}_k^-, k=1,\cdots,M$
							 \FOR{$q \gets 1$ to $iter$}
               \STATE  \hspace{-6mm} Update $\left\{ \boldsymbol{\alpha}_i \right\}_{i=1}^{N^+}$ for $\mathbf{x}_i \in \mathbf{B}_j^+$ using \eqref{eqn:alpha_plus_update}, \eqref{eqn:sf_alpha_pos}\\
							 \STATE \hspace{-6mm} Update $\left\{ \boldsymbol{\alpha}_i \right\}_{i=1}^{N^-}$ for $\mathbf{x}_i \in \mathbf{B}_j^-$ using \eqref{eqn:alpha_minus_update}
		           \ENDFOR
\STATE $l \gets l + 1$
\UNTIL{Convergence}
   \RETURN $\mathbf{D}$, $\left\{ \boldsymbol{\alpha}_i \right\}_{i=1}^N$\\
   
   \hspace{-5mm}\footnotesize{*DL-FUMI code can be found at: https://github.com/TigerSense/FUMI}
\end{algorithmic} 
\label{alg:gFUMI}
\end{algorithm}

The \emph{M-step} is performed by iteratively optimizing \eqref{eqn:E_gFUMI} for each of the desired parameters. The dictionary $\mathbf{D}$ is updated atom-by-atom using a block coordinate descent scheme \cite{bertsekas1999nonlinear, mairal2010online}.  The sparse weights, $\left\{ \boldsymbol{\alpha}_i \right\}_{i=1}^N$, are updated using an iterative shrinkage-thresholding algorithm \cite{figueiredo2003algorithm, daubechies2003iterative}.  For readability, the derivation of update equations are described in Sec. \ref{sec:gFUMI_update}.  The method is summarized in Alg. \ref{alg:gFUMI}. 
%



\section{Classification using Estimated Dictionary} \label{sec:6_classif}

Given  $\mathbf{D}$, a confidence that the $i^{th}$ instance is target can be computed using a ratio of the reconstruction errors given the target and background atoms, $\mathbf{D}$, vs. background atoms, $\mathbf{D}^-$:

\begin{equation}
c_{i}=\frac{\left\| \mathbf{x}_i - \boldsymbol{\alpha}_i^- \mathbf{D}^-\right\|^2}{\left\| \mathbf{x}_i - \boldsymbol{\alpha}_i \mathbf{D} \right\|^2},
\label{eqn:class}
\end{equation}
where $\boldsymbol{\alpha}_i^-$ are the sparse weights for the non-target  atoms for the $i^{th}$ instance. If the numerator has a large error and the denominator has a low error, then the target  atoms are needed to reconstruct instance $i$.  

\section{Experiments} \label{sec:7_experiments}
DL-FUMI is evaluated on two MIL AR Face data \cite{martinez1998ar, martinez2001pca} recognition problems and an MIL USPS hand-written digits \cite{lecun1989backpropagation, gader1996automatic} recognition problem. 
In all of our experiments, target atoms were initialized by computing mean of $T$ random subsets drawn from the union of all positive bags. $K$-means was applied to the union of all negative bags and the $M$ cluster centers were set as the initial background atoms.  

\subsection{AR Face Recognition} \label{sec:7_3_ARface}
The AR-face data set consists of frontal-pose images with 26 images/person (2 sessions, 13 per session) corresponding to different expressions, illuminations and occlusions. Pre-processed and cropped imagery of 50 male and 50 female subjects provided by Martinez and Kak \cite{martinez2001pca} was used. Each image was down-sampled to  $83\times 60$ pixels and the raw gray scale values were used as features.

For the first AR Face experiment, {sun-glasses} were the {target concept}. Specifically, 50 positive training bags of 10 instances each were created.  Each positive bag contained only two instances of randomly selected images of people wearing sun-glasses; the other eight were randomly chosen from images of people without sun-glasses. 50 negative bags were constructed by randomly selecting 10 instances per bag of images of individuals not wearing sun-glasses. Test data included all imagery that was not used for training.  

\begin{figure}
	\vspace{+3mm}
\begin{center}
\includegraphics[width=7.5cm]{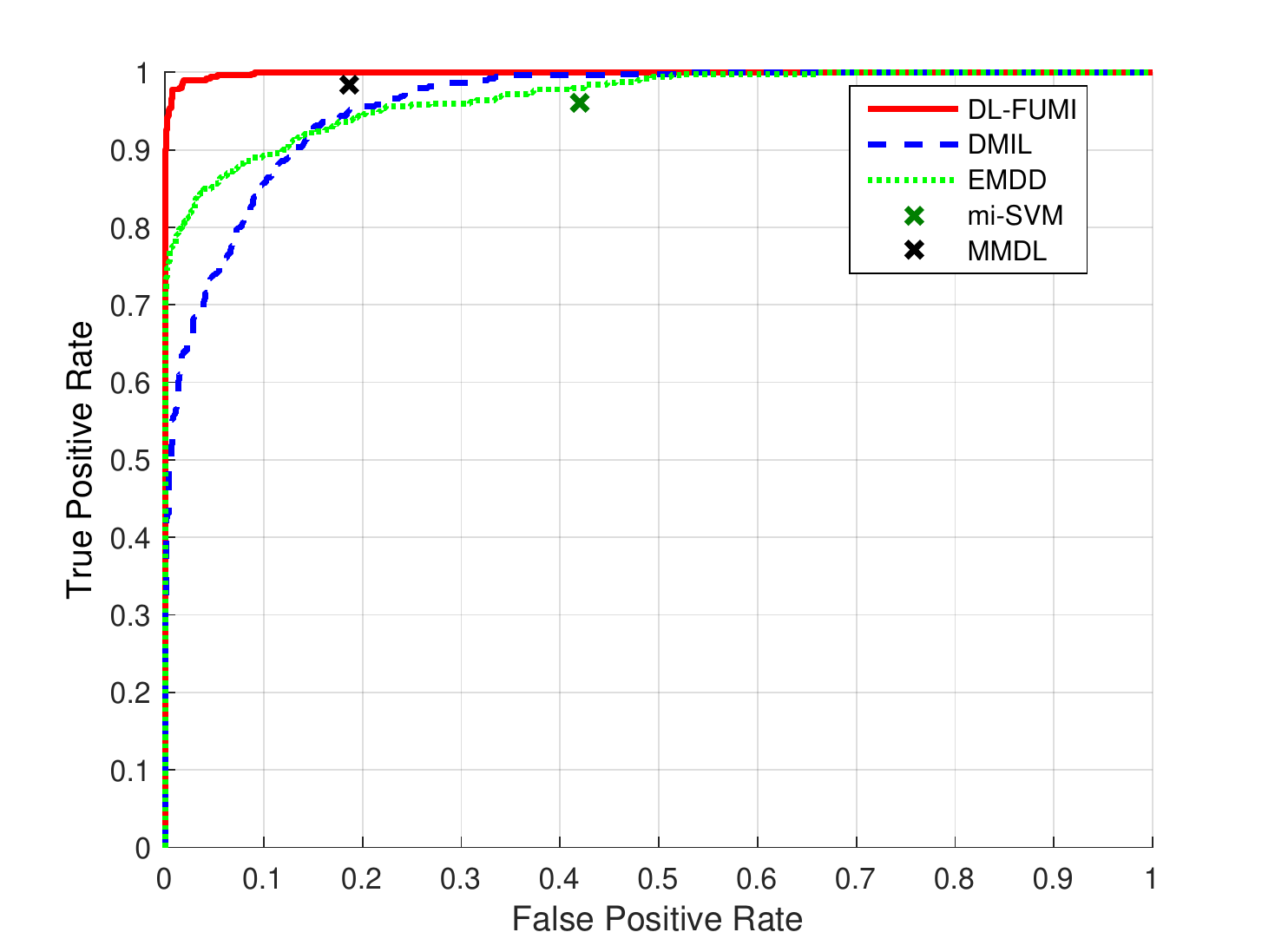} 
\caption{ROC analysis for sun-glasses detection using AR face database comparing DL-FUMI,  DMIL \cite{shrivastava2014dictionary} and EM-DD \cite{Zhang:2002} (code from \cite{zhou2003ensembles}).  True Positive Rate vs. False Positive Rate of mi-SVM \cite{andrews2002support}  and MMDL \cite{wang2013max} (code from author's website) are also plotted.}\label{fig:roc_sunglas_together}
\end{center}
\vspace{-5mm}
\end{figure}

\begin{figure}
	\begin{center}
		\subfigure[DL-FUMI]{
			\includegraphics[width=3cm]{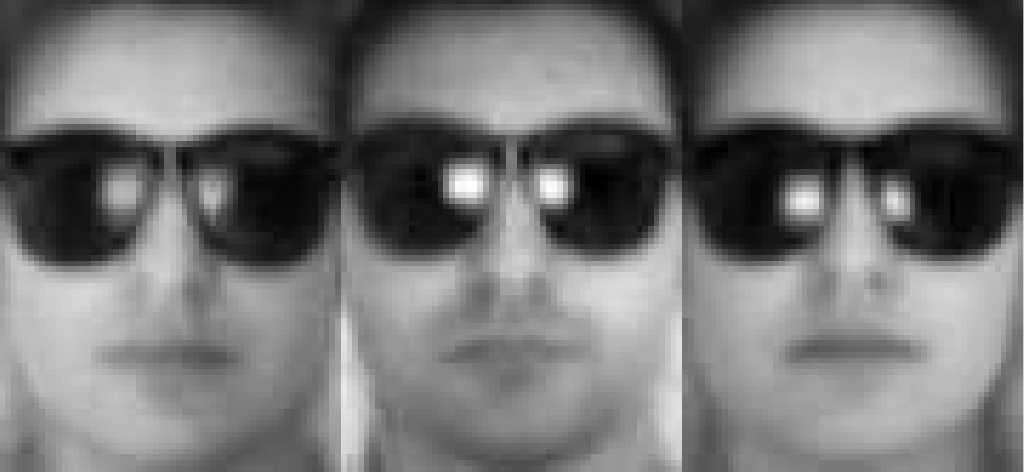} \label{fig:face_sunglas_tar_gFUMI}} 
		\subfigure[DMIL]{
			\includegraphics[width=3cm]{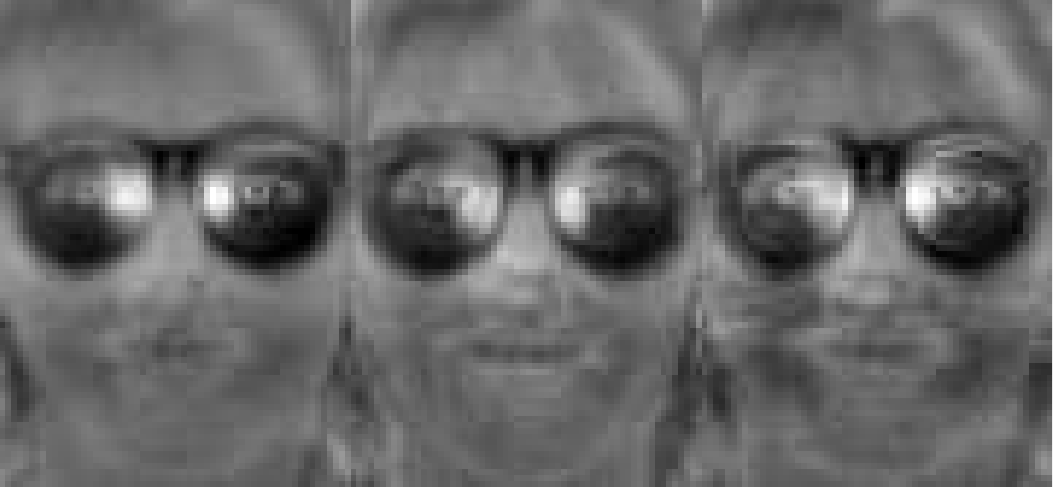} \label{fig:face_sunglas_tar_DMIL}} 
		\subfigure[EMDD]{
			\includegraphics[width=1.008cm]{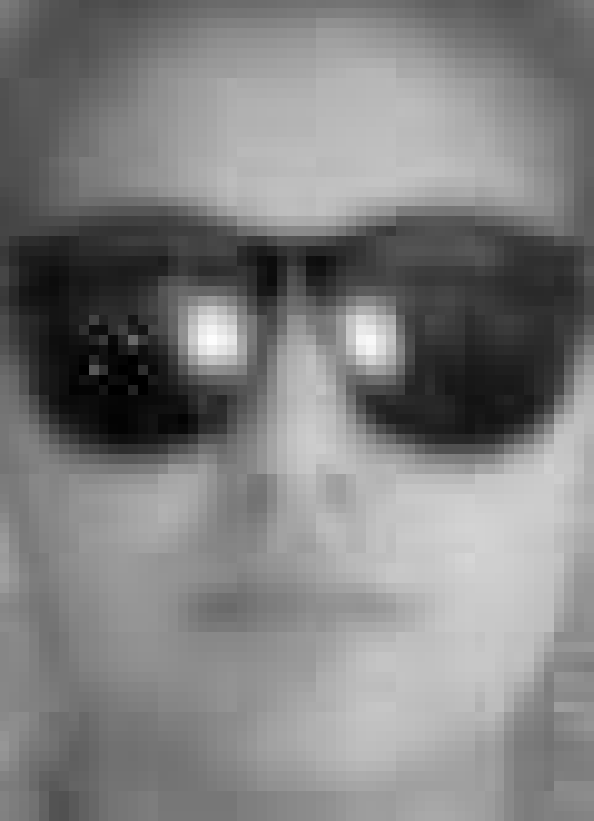} \label{fig:face_sunglas_tar_EMDD}}
		\caption{Plot of estimated dictionary atoms for sun-glasses. (a): DL-FUMI. (b): DMIL. (c): EM-DD.}\label{fig:AR_face_sungl_atoms}
	\end{center}
	\vspace{-6mm}
\end{figure}

\begin{figure}
	\begin{center}
		\subfigure[DL-FUMI]{
			\includegraphics[width=7cm]{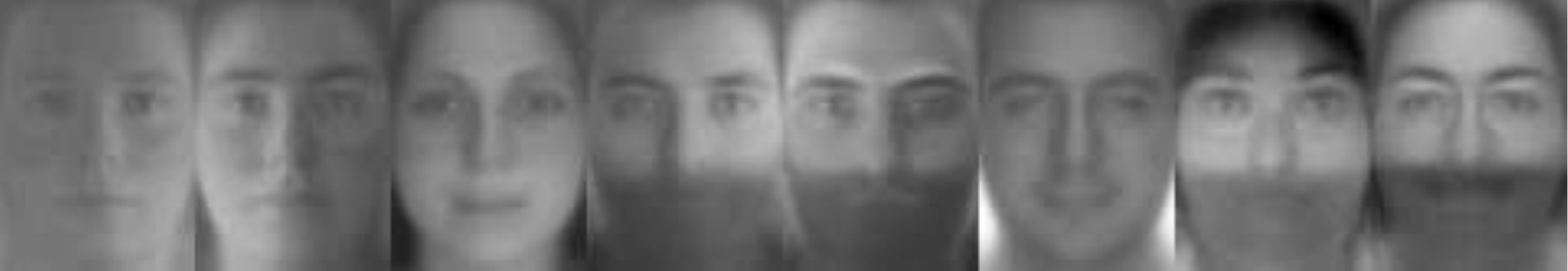} \label{fig:face_sunglas_nontar_gFUMI}} 
		\subfigure[DMIL]{
			\includegraphics[width=7cm]{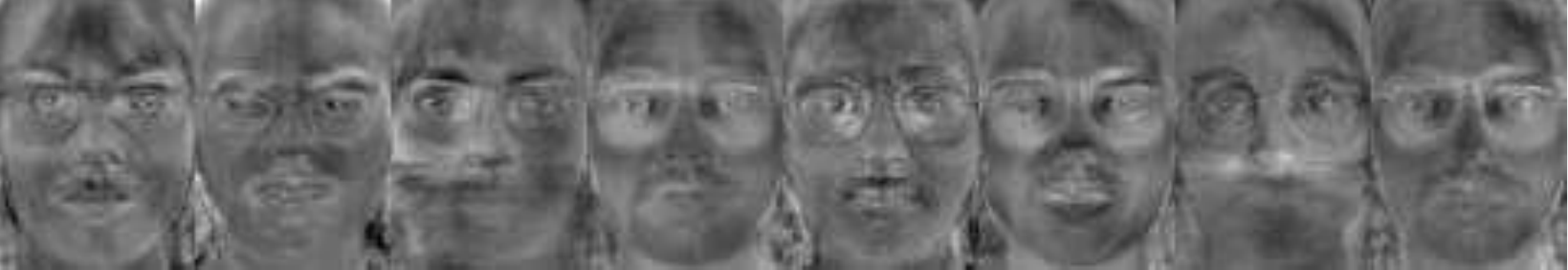} \label{fig:face_sunglas_nontar_DMIL}} 
		\caption{Plot of estimated dictionary atoms for background. (a): DL-FUMI. (b): DMIL.}\label{fig:AR_face_sungl_backgr_atoms}
	\end{center}
\end{figure}

The parameters for DL-FUMI for this experiment were set to $T=3$, $M=8$, $\Gamma=0.001$, $\beta=30$ and $\lambda=0.001$. After dictionary estimation, the target confidence was computed for each test instance following Sec. \ref{sec:6_classif}. Receiver operating characteristic (ROC) curve analysis was conducted. Fig. \ref{fig:roc_sunglas_together} shows one of the 10 ROCs obtained by DL-FUMI, DMIL and EM-DD where the TPR vs FPR obtained by mi-SVM and MMDL were also plotted. 
The average TPRs of DL-FUMI, EM-DD and DMIL over 10 runs are shown in Table \ref{tab:PD_AR_face_sunglas} at FPRs 1\%, 18.4\% and 41.9\%, where 18.4\% and 41.9\% are average FPRs by two classification algorithms MMDL and mi-SVM, respectively. 
Fig. \ref{fig:AR_face_sungl_atoms} and Fig. \ref{fig:AR_face_sungl_backgr_atoms} show estimated target and background atoms by DL-FUMI and comparison methods, respectively. To estimate the DMIL background atoms, we flipped the sign of positive and negative bags (\ie, swapped the target and background classes) and re-trained the dictionary.  This was done since, as stated in \cite{shrivastava2015gen}, DMIL does not learn a set of background atoms simultaneously.   
As shown, DL-FUMI target atoms are very discriminative and representative of the target class, \eg, there are male and female sun-glasses atoms and variation in light reflections. Finally, the overall dictionary set estimated by DL-FUMI is qualitatively more smooth which will help to reduce error in classification.

\begin{table}[!htb] 
\vspace{+5mm}
\begin{center}
\caption{Average TPR at FPRs over 10 runs}\label{tab:PD_AR_face_sunglas}
\begin{tabular}{|c|c|c|c|c|}
\hline
Algorithm  & \makecell{TPR(\%)\\FPR=1\%} & \makecell{TPR(\%)\\FPR=18.4\%} &\makecell{TPR(\%)\\FPR=41.9\%}\\
\hline
\hline
mi-SVM \cite{andrews2002support}& - & - & 96.2 \\
\hline
EM-DD \cite{Zhang:2002}& 78.2 & 93.8  &98.1\\
\hline
MMDL \cite{wang2013max} &  -  & 98.0 & -\\
\hline
DMIL \cite{shrivastava2014dictionary} &60.2 &95.2 &99.5\\
\hline
\textbf{DL-FUMI} & 97.5 & 100 & 100\\
\hline
\end{tabular}
\end{center}
\vspace{-2mm}
\end{table}

For the second AR Face experiment, Woman No. 10 was selected as the positive target class. Two positive training bags containing 50 instances each were created. The first positive bag contained 6 images from {Woman} No. 10 set 1 and the second positive bag contained the remaining 7 images from {Woman} No. 10 set 1, and the rest of the instances in each positive bag were randomly selected from other individuals.  Three negative bags with 200 instances per bag were constructed by randomly selecting images from the data set excluding those from {Woman} No. 10. Given this, there are only 13 positive training instances and 687 negative training instances. This is a more difficult problem than {sun-glass} detection. The test data contained the 13 images of Woman No. 10  from set 2 and 100 images randomly selected from images that are not {Woman} No. 10. There is no overlap between the training and testing data.

\begin{figure}
\vspace{+3mm}
\begin{center}
\includegraphics[width=7.5cm]{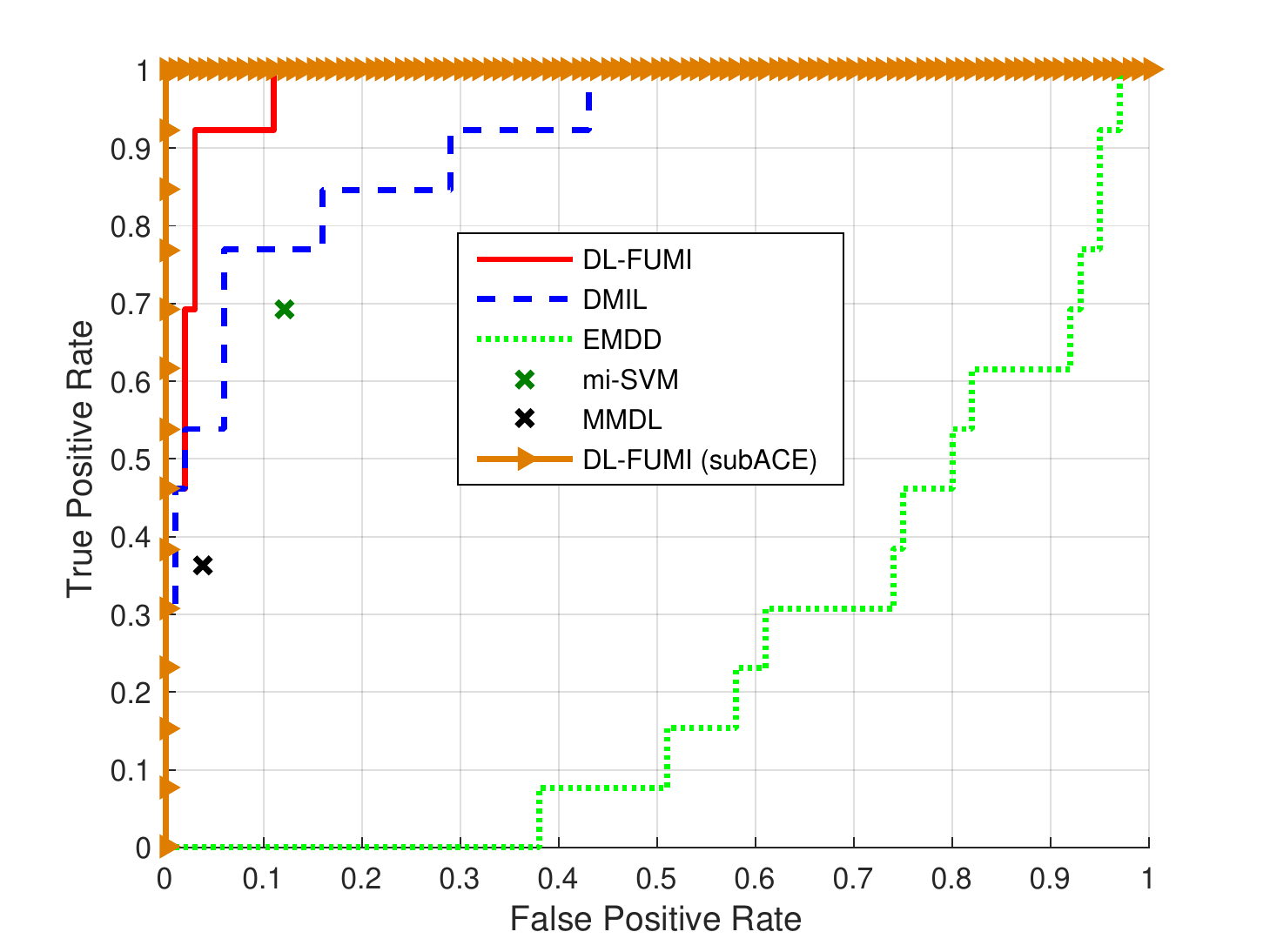} 
\caption{{Woman} No. 10 detection on AR face database}\label{fig:roc_wm10_together}
\end{center}
\vspace{-3mm}
\end{figure}

\begin{figure}
	\begin{center}
		\subfigure[]{
			\includegraphics[width=3cm]{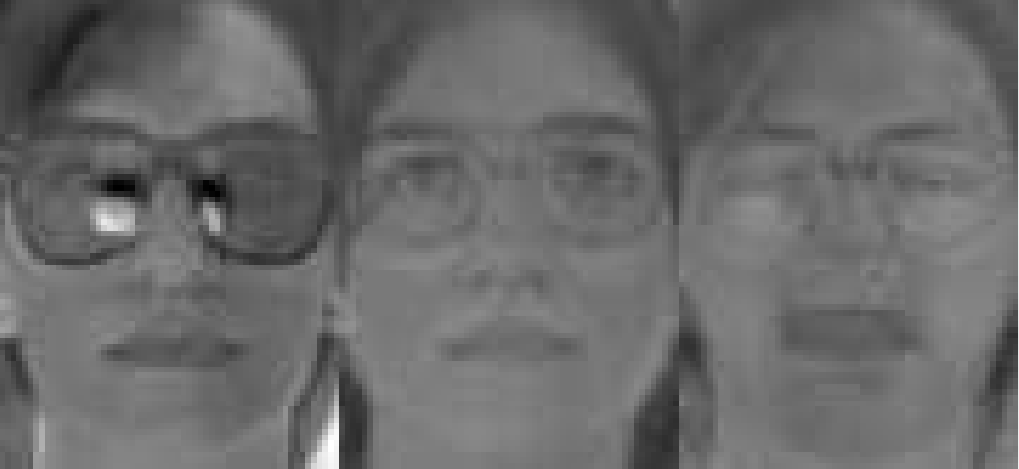} \label{fig:face_wm10_tar_gFUMI}} 
		\subfigure[]{
			\includegraphics[width=3cm]{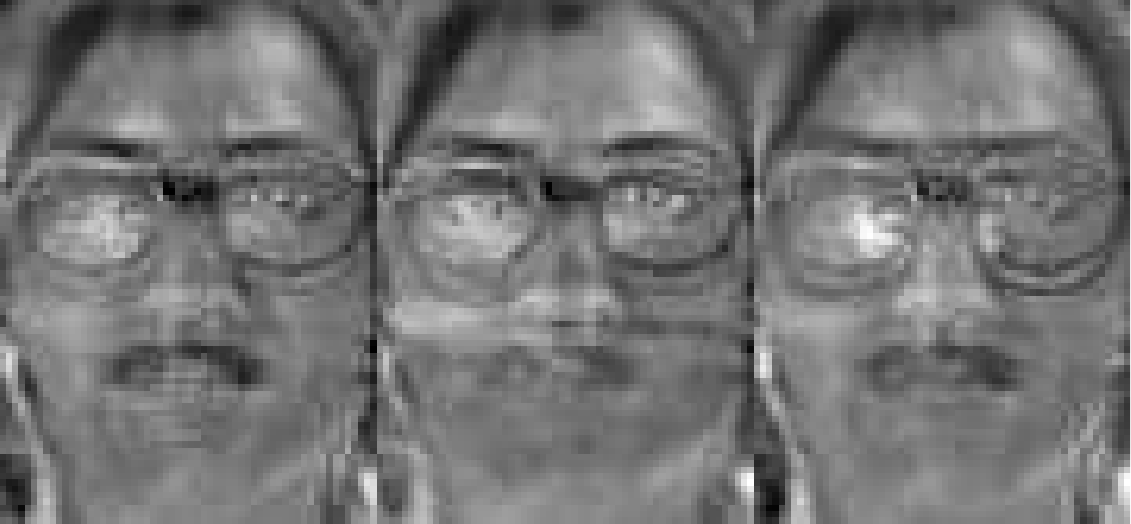} \label{fig:face_wm10_tar_DMIL}} 
		\subfigure[]{
			\includegraphics[width=1.008cm]{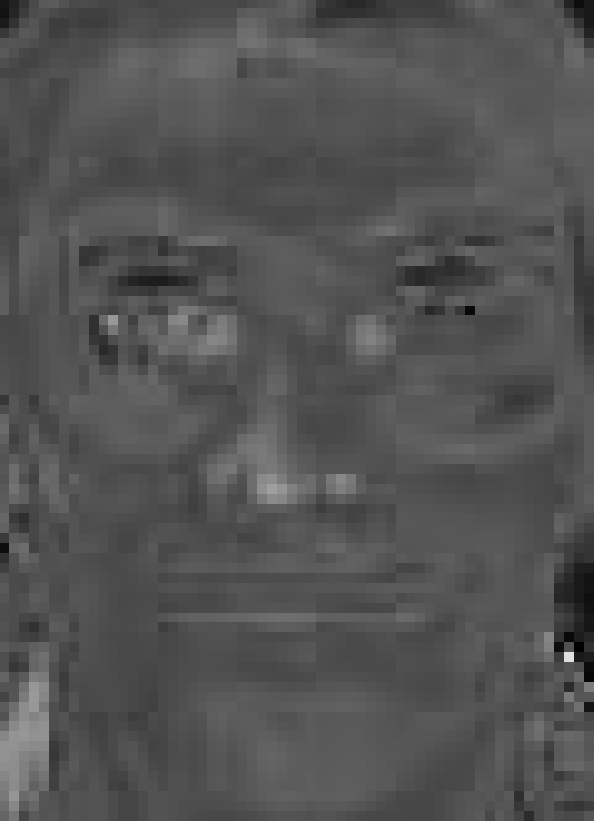} \label{fig:face_wm10_tar_EMDD}} 
		\caption{Plot of estimated dictionary atoms for {Woman} No. 10. (a): DL-FUMI. (b): DMIL. (c): EM-DD}\label{fig:AR_face_wm10_tar_atoms}
	\end{center}
\vspace{-5mm}
\end{figure}

\begin{figure}
	\vspace{+3.5mm}
	\begin{center}
		\subfigure[]{
			\includegraphics[width=7.5cm]{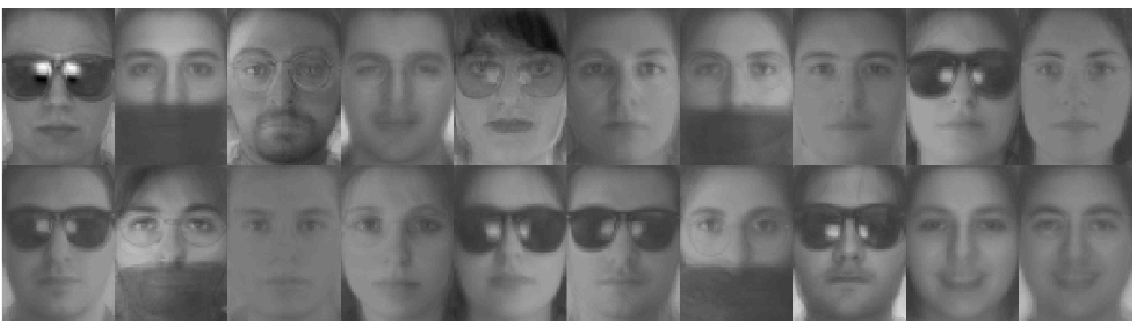} \label{fig:face_wm10_backgr_gFUMI}}
		\subfigure[]{
			\includegraphics[width=7.58cm]{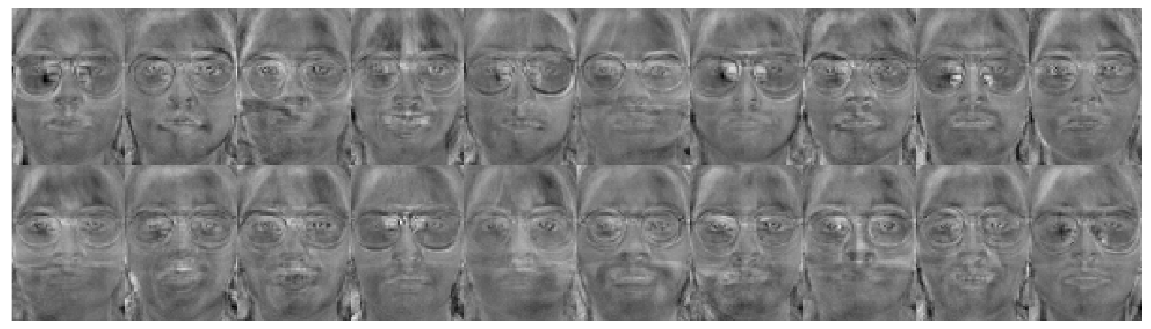} \label{fig:face_wm10_backgr_DMIL}}  
		\caption{Plot of estimated background dictionary atoms for {Woman} No. 10.}\label{fig:AR_face_wm10_backgr_atoms}
	\end{center}
	\vspace{-6mm}
\end{figure}

The parameters used in DL-FUMI for this experiment were $T=3$, $M=20$, $\Gamma=0.001$, $\beta=50$ and $\lambda=0.001$.  One of 10 ROCs is shown in Fig. \ref{fig:roc_wm10_together}. The average TPRs of DL-FUMI, EM-DD and DMIL are shown in Table \ref{tab:PD_AR_face_wm10} at FPRs 2.9\%, 5\% and 12.0\%, where 2.9\% and 12.0\% are average FPRs by two classification algorithms MMDL and mi-SVM, respectively. Table \ref{tab:PD_AR_face_wm10} and Fig. \ref{fig:roc_wm10_together} clearly show that DL-FUMI outperforms all the comparison algorithms. In order to further show that the estimated target dictionary by DL-FUMI is effective at characterizing the target class, the subspace adaptive cosine estimator (subACE) target detection algorithm \cite{kraut:2001} was applied for detection using the target dictionary estimated by DL-FUMI directly. One of the subACE ROCs using the DL-FUMI dictionary shows a 100\% TPR with 0\% FPR in Fig. \ref{fig:roc_wm10_together}.  Since subACE is a target detection algorithm that relies on target signatures that encompass the distinguishing characteristics of a target class, this further emphasizes that target dictionary estimated by DL-FUMI is highly representative of the target class. Fig. \ref{fig:face_wm10_tar_gFUMI} - \ref{fig:face_wm10_tar_EMDD} show the target atoms estimated by DL-FUMI, DMIL and EM-DD for {Woman} No. 10, where it can be seen that the target dictionary atoms estimated by DL-FUMI are more discriminative as it captures different distinct features of the positive class (different occlusions, expressions, \etc). Fig. \ref{fig:face_wm10_backgr_gFUMI} and \ref{fig:face_wm10_backgr_DMIL} show the background atoms estimated for Women No. 10 and it can be seen that the background dictionary estimated by DL-FUMI has better representative quality.
\vspace{-3mm}

\begin{table}[!htb] 
\begin{center}
\caption{Average TPR at FPRs over 10 runs}\label{tab:PD_AR_face_wm10}
\begin{tabular}{|c|c|c|c|c|}
\hline
Algorithm & \makecell{TPR(\%)\\FPR=2.9\%} & \makecell{TPR(\%)\\FPR=5\%}  &\makecell{TPR(\%)\\FPR=12.0\%}\\
\hline
\hline
mi-SVM \cite{andrews2002support}& - & - & 69.2 \\
\hline
EM-DD \cite{Zhang:2002}& 0 & 0 & 0 \\
\hline
MMDL \cite{wang2013max}& 31.5 & -  & - \\
\hline
DMIL \cite{shrivastava2014dictionary}& 54.5 &69.1 & 78.8 \\
\hline
\textbf{DL-FUMI} (using \eqref{eqn:class}) &78.9 & 91.5 &95.5\\
\hline
\textbf{DL-FUMI} (subACE) & 94.6 &100 &100\\
\hline
\end{tabular}
\end{center}
\vspace{-3mm}
\end{table}

\subsection{USPS Digit Classification} \label{sec:7_2_usps}
DL-FUMI is further evaluated on  a multi-class classification task given the USPS\footnotemark[2] data set. The USPS data set contains 9298 images of handwritten digits from 0 to 9. Each image is $16\times16$ in size. The raw gray-level pixel values are used as features in this  experiment. The training and testing data partitions in this paper mimics the experimental set-up in \cite{shrivastava2015gen}. Specifically, for each class $c$, 50 positive training bags were generated.  Each bag contains 4 instances in total and only one comes from true $c^{th}$ positive class and the other three instances are randomly chosen from other classes. 50 negatively labeled bags were also constructed by randomly selecting 50 instances per bag from classes other than $c$. The testing data contains 2000 samples in total, 200 from each class.

\begin{figure}
 	\vspace{+4mm}
 	\begin{center}
 		\subfigure[]{
 			\includegraphics[width=1.505cm]{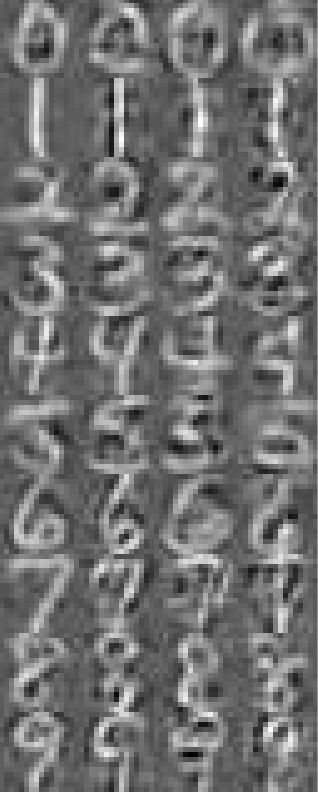} \label{fig:usps_tar_atoms}} 
 		\subfigure[]{   
 			\includegraphics[width=5.65cm]{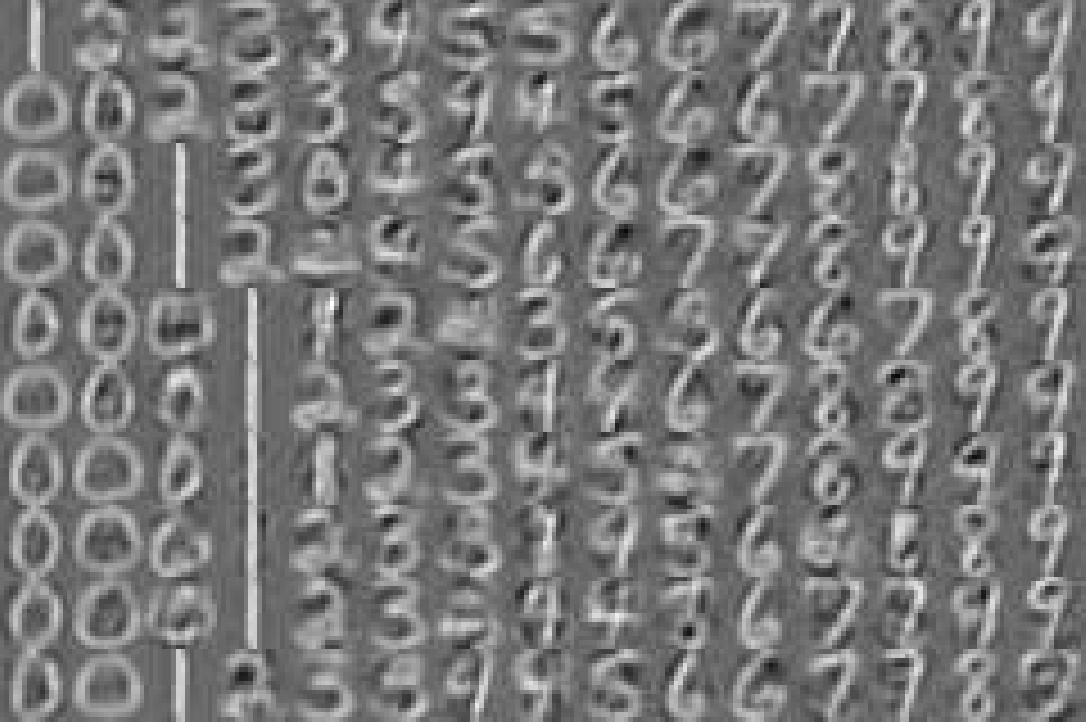} \label{fig:usps_nontar_atoms}} 
 		\caption{USPS dictionary atoms estimated by DL-FUMI. (a): Target atoms. (b): Non-target atoms.}\label{fig:usps_dict_plot}
 	\end{center}
 	\vspace{-3mm}
 \end{figure}

In this experiment, the parameters used were $T=4$, $M=15$, $\Gamma=0.1$, $\beta=25$ and $\lambda=0.001$.  For instance level classification, the approach described in Sec. \ref{sec:6_classif} was applied given a dictionary estimated given each class as the target class.  Then, the final class label, for multi-class classification, was assigned by selecting the class with the largest confidence value computed in \eqref{eqn:class}. The classification results of DL-FUMI and comparison algorithms are listed in Table \ref{tab:usps}, where results for GD-MIL are as reported in \cite{shrivastava2015gen}. Table \ref{tab:usps} shows that DL-FUMI outperforms two multiple instance dictionary learning methods, GD-MIL and MMDL, and two MIL methods mi-SVM and EM-DD. Fig. \ref{fig:usps_tar_atoms} and \ref{fig:usps_nontar_atoms} show the estimated DL-FUMI target and non-target dictionary atoms, in these we can see that DL-FUMI is able to learn a set of discriminative target dictionary as well as characteristic background dictionary (\ie, each target dictionary atom looks like the target digit and the background dictionary atoms look like all the other digits). 

\begin{figure}
 	\begin{center}
 		\subfigure[]{
 			\includegraphics[width=1.45cm]{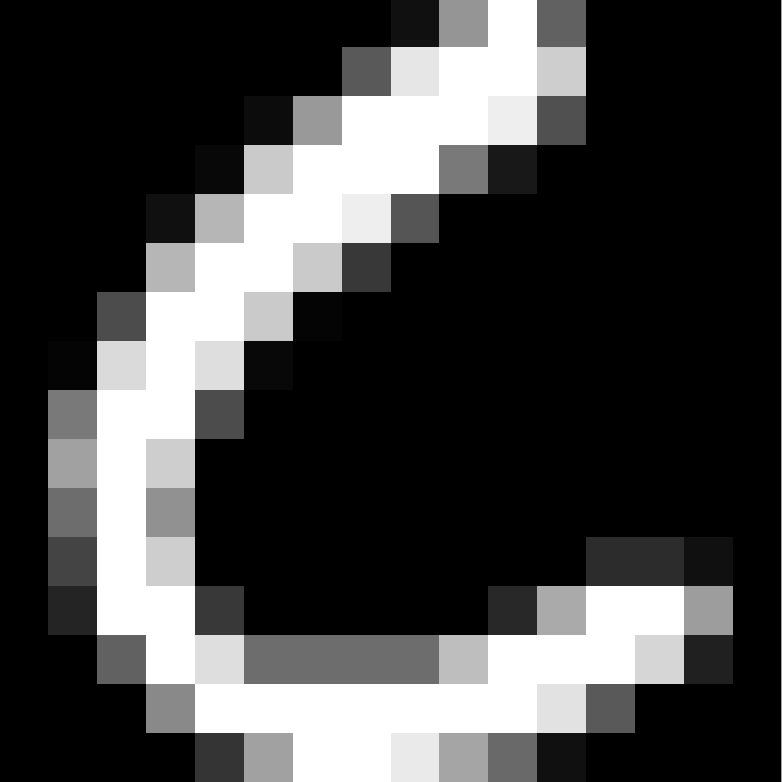} \label{fig:usps_miss_0}} 
 		\subfigure[]{
 			\includegraphics[width=1.45cm]{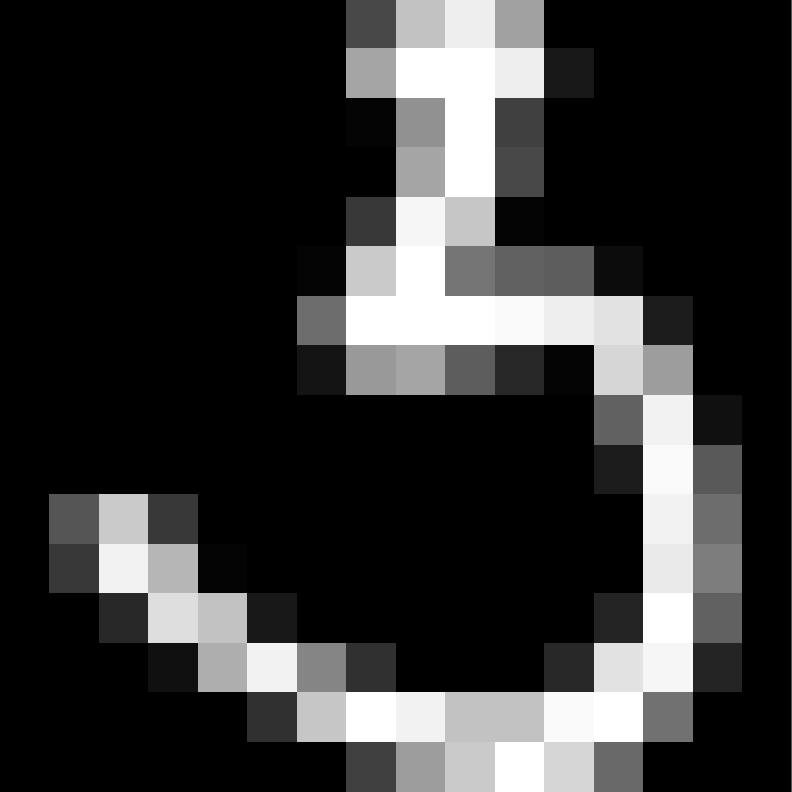} \label{fig:usps_miss_5}} 
 		\subfigure[]{
 			\includegraphics[width=1.45cm]{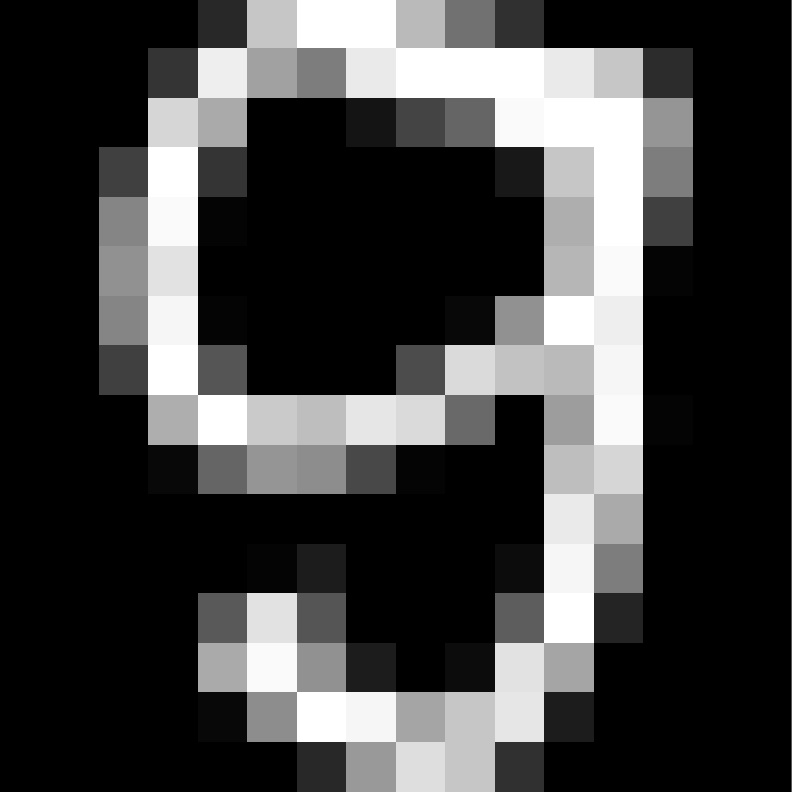} \label{fig:usps_miss_9}} 
 		\subfigure[]{
 			\includegraphics[width=1.45cm]{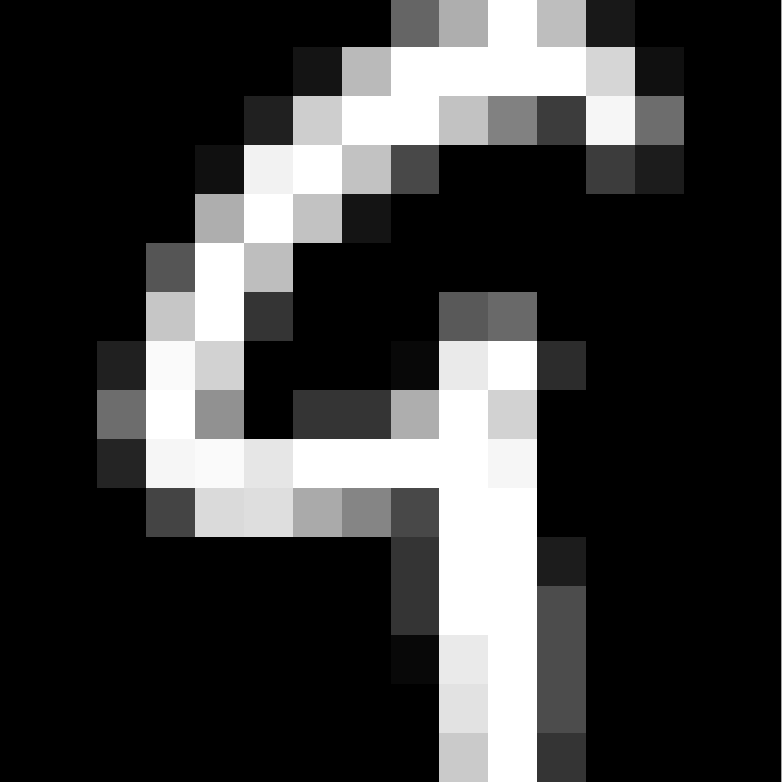} \label{fig:usps_miss_9_1}} 
 		\subfigure[]{
 			\includegraphics[width=1.45cm]{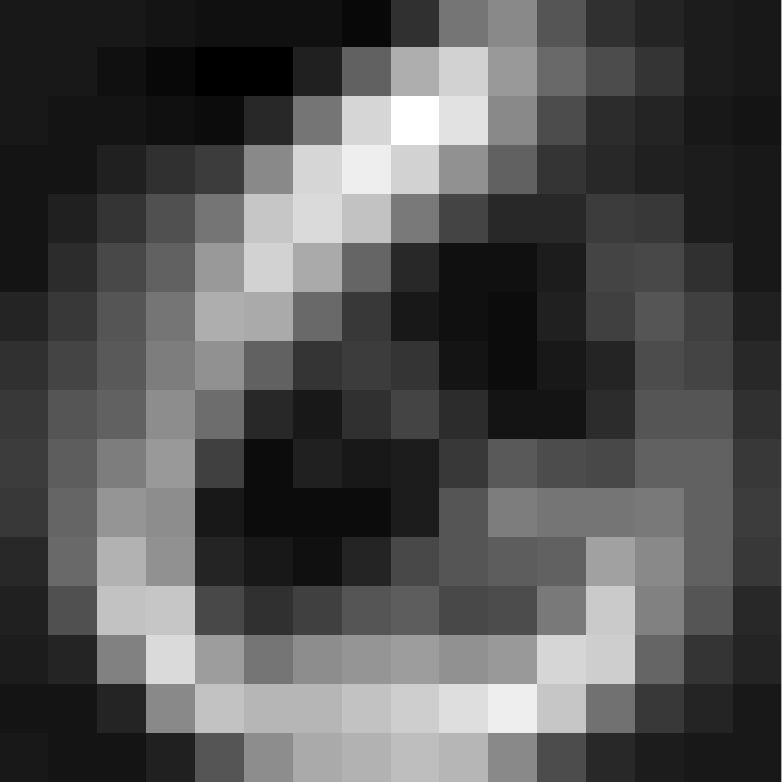} \label{fig:usps_recon_0}} 
 		\subfigure[]{
 			\includegraphics[width=1.45cm]{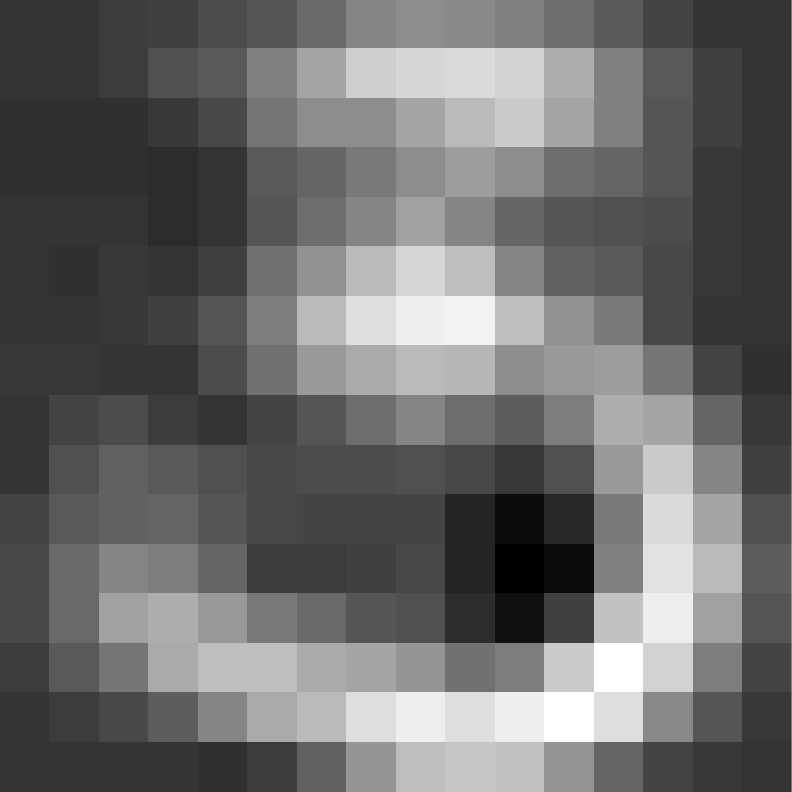} \label{fig:usps_recon_5}} 
 		\subfigure[]{
 			\includegraphics[width=1.45cm]{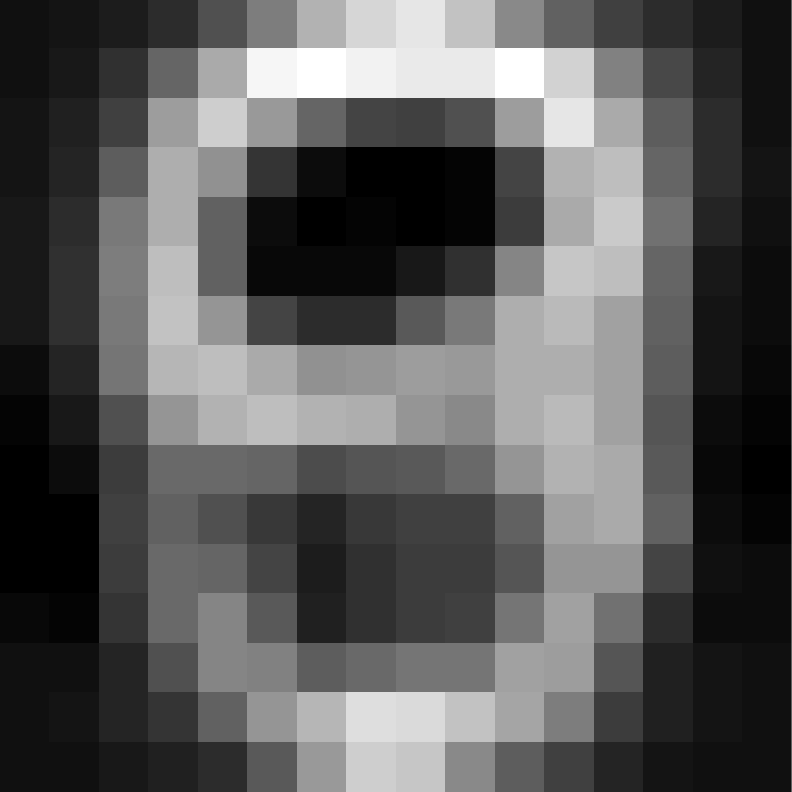} \label{fig:usps_recon_9}} 
 		\subfigure[]{
 			\includegraphics[width=1.45cm]{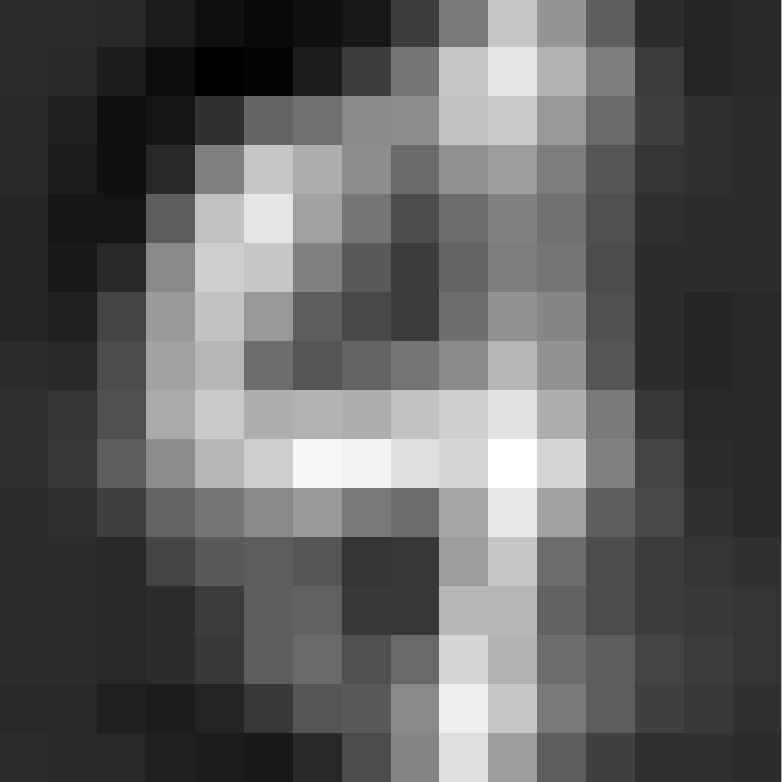} \label{fig:usps_recon_9_1}} 
 		\caption{Examples of misclassified images by DL-FUMI. (a): true class 0 misclassified to 6. (b): true class 5 misclassified to 3. (c): true class 9 misclassified to 8. (d): true class 9 misclassified to 4. (e)-(h): corresponding reconstructed images by DL-FUMI}\label{fig:usps_miss_plot}
 	\end{center}
 	\vspace{-2mm}
 \end{figure}

To get insight into classification errors, Fig. \ref{fig:usps_miss_0} - \ref{fig:usps_miss_9_1} show several examples of randomly selected misclassified instances and Fig. \ref{fig:usps_recon_0} - \ref{fig:usps_recon_9_1} show the reconstructed images by DL-FUMI. For example, Fig. \ref{fig:usps_miss_0} has a true class label of 0, but was misclassified to 6; the reconstructed image is shown in Fig. \ref{fig:usps_recon_0}. As can be seen, this data point appears to be very similar to the digit 6 and is even difficult for a human to correctly recognize. Similarly, Fig. \ref{fig:usps_miss_5} - \ref{fig:usps_miss_9_1} show the other three images and Fig. \ref{fig:usps_recon_5} - \ref{fig:usps_recon_9_1} show the corresponding reconstructed images, respectively.
\footnotetext[2]{Database at: http://www-i6.informatik.rwth-aachen.de/~keysers/usps.html}

\begin{table}[!htb] 
\vspace{-3mm}
\begin{center}
\caption{Classification accuracies on USPS data set}\label{tab:usps}
\begin{tabular}{|c|c|c|c|}
\hline
Alg.  & Acc.($\%$) & Alg.  & Acc.($\%$)\\
\hline
\hline
mi-SVM \cite{andrews2002support}& 81.1 & GD-MIL\footnotemark[3] \cite{shrivastava2015gen} & \underline{83.4} \\
\hline
EM-DD \cite{Zhang:2002}& 76.55 & MMDL \cite{wang2013max} &80.8 \\
\hline
\textbf{DL-FUMI} & \textbf{86.5} & &\\
\hline
\end{tabular}
\end{center}
\vspace{-5mm}
\end{table}

 \footnotetext[3]{Results as stated in the literature \cite{shrivastava2015gen}.}

\begin{figure*}[!hbt]
	\begin{footnotesize}
		\begin{eqnarray}
		\mathbf{d}_k^-&=&\left\{\sum_{i=1}^{N^+} \left[P(z_i=1)\psi\alpha_{ik}(\mathbf{x}_i - \sum_{t=1}^T\alpha_{it}\mathbf{d}_t^+-\sum_{l=1,l\neq k}^M\alpha_{il}\mathbf{d}_l^-)+ P(z_i=0)\psi\alpha_{ik}(\mathbf{x}_i - \sum_{l=1,l\neq k}^M\alpha_{il}\mathbf{d}_l^-)\right]+\sum_{i=1}^{N^-} \left[\alpha_{ik}(\mathbf{x}_i - \sum_{l=1,l\neq k}^M\alpha_{il}\mathbf{d}_l^-)\right]\right. \nonumber \\
		&&\left.-\Gamma\sum_{t=1}^T\cos\theta_{kt}\mathbf{d}_{t^{old}}^+\right\}\left\{\sum_{i=1}^{N+}\psi\alpha_{ik}^2+\sum_{i=1}^{N-}\alpha_{ik}^2\right\}^{-1} 
		\label{eqn:update_dk}
		\end{eqnarray}
		\hrulefill
	\end{footnotesize}
\vspace{-3mm}
\end{figure*}

\section{Conclusion} \label{sec:8_conclusion}
In this paper, a multiple-instance dictionary learning algorithm, DL-FUMI, is proposed. DL-FUMI leverages the \textit{shared} information between positive and negative classes to improve the discriminative ability of the estimated target atoms. Experimental results show that the estimated DL-FUMI target atoms  provide a good representation of the positive class and  improves target detection and classification performance over comparison methods. 
\section{Derivation of DL-FUMI update equations}
\label{sec:gFUMI_update}
This section provides a derivation of DL-FUMI update equations. When updating the dictionary $\mathbf{D}$, the sparse weights $\left\{\boldsymbol{\alpha}_i\right\}_{i=1}^{N}$ are held fixed.  
To update one of the atoms in $\mathbf{D}$, \eqref{eqn:E_gFUMI} is minimized with respect to the corresponding atom while keeping all other atoms constant. The resulting update equations for $\mathbf{d}_{t}^+$ and $\mathbf{d}_{k}^-$  are shown in   \eqref{eqn:update_dt} and \eqref{eqn:update_dk}.  

\begin{footnotesize}
\begin{equation}
\mathbf{d}_t^+=\frac{\sum_{i=1}^{N^+} \left[P(z_i=1)\alpha_{it}(\mathbf{x}_i - \sum_{l=1,l\neq t}^T\alpha_{il}\mathbf{d}_l^+-\sum_{k=1}^M\alpha_{ik}\mathbf{d}_k^-)\right]}{\sum_{i=1}^{N+}\left[P(z_i=1)\alpha_{it}^2\right]} 
\label{eqn:update_dt}
\end{equation}
\end{footnotesize}
Note,  $P(z_i | \mathbf{x}_i, \boldsymbol{\theta}^{(t-1))}$ is denoted as $P(z_i)$ for simplicity.

When updating the sparse weights,  $\left\{\boldsymbol{\alpha}_i\right\}_{i=1}^N$, it should be noted that the sparse weight vector $\boldsymbol{\alpha}_i$ for instance $\mathbf{x}_i$ is not dependent on any other instances.
%
%

The gradient with respect to $\boldsymbol{\alpha}_i$ without considering the $l_1$ penalty term is: 
\begin{small}
\begin{eqnarray}
&&\frac{\partial F^+}{\partial \boldsymbol{\alpha}_i}=-\begin{bmatrix}P(z_i=1)\mathbf{D}^+ & \mathbf{D}^-\end{bmatrix}^T\mathbf{x}_i+\left( P(z_i=1)\mathbf{D}^T\mathbf{D}\right.\nonumber \\ 
&&\left.+P(z_i=0)\begin{bmatrix}\mathbf{0}_{d \times T} & \mathbf{D}^-\end{bmatrix}^T\begin{bmatrix}\mathbf{0}_{d \times T} & \mathbf{D}^-\end{bmatrix} \right)\boldsymbol{\alpha}_i. 
\label{eqn:gradient_alpha_plus}
\end{eqnarray}
\end{small}
Then $\boldsymbol{\alpha}_i$ at $l^{th}$ iteration can be updated using gradient descent,

\begin{equation}
\boldsymbol{\alpha}_i^{l}=\boldsymbol{\alpha}_i^{l-1}-\eta_i\frac{\partial F^+}{\partial \boldsymbol{\alpha}_i},
\label{eqn:alpha_plus_update}
\end{equation} 
followed by a soft-thresholding:
\begin{eqnarray}
\left\{ \begin{array}{l}
\boldsymbol{\alpha}_{i}^{l+}=S_{\lambda P(z_i=1)}\left(\boldsymbol{\alpha}_i^{l+}\right) \\
\boldsymbol{\alpha}_{i}^{l-}=S_{\lambda}\left(\boldsymbol{\alpha}_i^{l-}\right)
\label{eqn:sf_alpha_pos} 
\end{array}\right.,
\end{eqnarray}
s.t. $S_{\lambda}\left(\mathbf{x}[i]\right)=sign(\mathbf{x}[i])\max(|\mathbf{x}[i]|-\lambda,0)$, $i=1,...,d$.

Following a similar proof to that in \cite{facchinei2007finite}, when \begin{scriptsize}$\eta_i\in\left(0, \left(\lambda_{max}\left(P(z_i=0)\left[\mathbf{0}_{d \times T} \;  \mathbf{D}^-\right]^T\left[\mathbf{0}_{d \times T} \; \mathbf{D}^-\right]+P(z_i=1)\mathbf{D}^T\mathbf{D}\right)\right)^{-1}\right)$\end{scriptsize}, the update of $\boldsymbol{\alpha}_i$ using a gradient descent method with step length $\eta_i$ monotonically decreases the value of the objective function, where $\lambda_{max}(\mathbf{A})$ denotes the maximum eigenvalue of $\mathbf{A}$. For simplicity,  $\eta$ was set as $\eta=\frac{1}{\lambda_{max}\left( \mathbf{D}^T\mathbf{D}\right)}$ for all $\boldsymbol{\alpha}_i$, $\mathbf{x}_i\in\mathbf{B}_j^+$. 
%
%
%
%

A similar update can be used for points from negative bags.  The resulting update equation for negative points is:

\begin{small}
\begin{equation}
\boldsymbol{\alpha}_i^l=S_{\lambda}\left(\boldsymbol{\alpha}^{l-1}_i+\frac{1}{\lambda_{max}\left(\mathbf{D}^{-T}\mathbf{D}^-\right)}\left(\mathbf{D}^{-T}(\mathbf{x}_i-\mathbf{D}^-\boldsymbol{\alpha}^{l-1}_i)\right)\right)
\label{eqn:alpha_minus_update}
\end{equation}
\end{small}

The sparse weights corresponding to target dictionary atoms are set to 0 for all points in negative bags.




\section*{Acknowledgment}
{This material is based upon work supported by the National Science Foundation under Grant No. IIS-1350078 - CAREER: Supervised Learning for Incomplete and Uncertain Data.}
{
\bibliographystyle{IEEETran}
\bibliography{gFUMI_CVPR_Ref}

\begin{thebibliography}{10}
\providecommand{\url}[1]{#1}
\csname url@samestyle\endcsname
\providecommand{\newblock}{\relax}
\providecommand{\bibinfo}[2]{#2}
\providecommand{\BIBentrySTDinterwordspacing}{\spaceskip=0pt\relax}
\providecommand{\BIBentryALTinterwordstretchfactor}{4}
\providecommand{\BIBentryALTinterwordspacing}{\spaceskip=\fontdimen2\font plus
\BIBentryALTinterwordstretchfactor\fontdimen3\font minus
  \fontdimen4\font\relax}
\providecommand{\BIBforeignlanguage}[2]{{%
\expandafter\ifx\csname l@#1\endcsname\relax
\typeout{** WARNING: IEEEtran.bst: No hyphenation pattern has been}%
\typeout{** loaded for the language `#1'. Using the pattern for}%
\typeout{** the default language instead.}%
\else
\language=\csname l@#1\endcsname
\fi
#2}}
\providecommand{\BIBdecl}{\relax}
\BIBdecl

\bibitem{mairal2014sparsemodeling}
J.~Mairal, F.~Bach, and J.~Ponce, ``Sparse modeling for image and vision
  processing,'' \emph{Found. and Trends in Comput. Graph. and Vision}, vol.~8,
  no. 2-3, pp. 85--283, 2014.

\bibitem{Dietterich:1997}
T.~G. Dietterich, R.~H. Lathrop, and T.~Lozano-Perez, ``Solving the
  multiple-instance problem with axis-parallel rectangles,'' \emph{Artificial
  Intell.}, vol.~89, no. 1-2, pp. 31--17, 1997.

\bibitem{andrews2002support}
S.~Andrews, I.~Tsochantaridis, and T.~Hofmann, ``Support vector machines for
  multiple-instance learning,'' in \emph{Advances in Neural Inf. Process.
  Syst.}, 2002, pp. 561--568.

\bibitem{chen2006miles}
Y.~Chen, J.~Bi, and J.~Z. Wang, ``{MILES}: Multiple-instance learning via
  embedded instance selection,'' \emph{IEEE Trans. Pattern Anal. Mach. Intell},
  vol.~28, no.~12, pp. 1931--1947, 2006.

\bibitem{Maron:1998}
O.~Maron and T.~Lozano-Perez, ``A framework for multiple-instance learning.''
  in \emph{Advances in Neural Inf. Process. Syst.}, vol.~10, 1998.

\bibitem{Zhang:2002}
Q.~Zhang and S.~Goldman, ``{EM-DD: An improved multiple-instance learning
  technique},'' in \emph{Advances in Neural Inf. Process. Syst.}, vol.~2.\hskip
  1em plus 0.5em minus 0.4em\relax MIT; 1998, 2002, pp. 1073--1080.

\bibitem{Zare:2015fumi}
C.~Jiao and A.~Zare, ``Functions of multiple instances for learning target
  signatures,'' \emph{IEEE Trans. on Geosci. Remote Sens.}, vol.~53, no.~8, pp.
  4670 -- 4686, 2015.

\bibitem{mallat1993matching}
S.~G. Mallat and Z.~Zhang, ``Matching pursuits with time-frequency
  dictionaries,'' \emph{IEEE Trans. Signal Process.}, vol.~41, no.~12, pp.
  3397--3415, 1993.

\bibitem{donoho2006compressed}
D.~L. Donoho, ``Compressed sensing,'' \emph{IEEE Trans. Inf. Theory}, vol.~52,
  no.~4, pp. 1289--1306, 2006.

\bibitem{mairal2012task}
J.~Mairal, F.~Bach, and J.~Ponce, ``Task-driven dictionary learning,''
  \emph{IEEE Trans. Pattern Anal. Mach. Intell.}, vol.~34, no.~4, pp. 791--804,
  2012.

\bibitem{jiang2013label}
Z.~Jiang, Z.~Lin, and L.~S. Davis, ``Label consistent {K-SVD}: Learning a
  discriminative dictionary for recognition,'' \emph{IEEE Trans. Pattern Anal.
  Mach. Intell.}, vol.~35, no.~11, pp. 2651--2664, 2013.

\bibitem{wang2013max}
X.~Wang, B.~Wang, X.~Bai, W.~Liu, and Z.~Tu, ``Max-margin multiple-instance
  dictionary learning,'' in \emph{Int. Conf. On Mach. Learning}, 2013, pp.
  846--854.

\bibitem{shrivastava2014dictionary}
A.~Shrivastava, J.~K. Pillai, V.~M. Patel, and R.~Chellappa, ``Dictionary-based
  multiple instance learning,'' in \emph{IEEE Int. Conf. on Image Process.},
  2014, pp. 160--164.

\bibitem{shrivastava2015gen}
A.~Shrivastava, V.~M. Patel, j.~K. Pillai, and R.~Chellappa, ``Generalized
  dictionaries for multiple instance learning,'' \emph{Int. J. of Comput.
  Vision}, vol. 114, no.~2, pp. 288--305, Septmber 2015.

\bibitem{ramirez2010classification}
I.~Ramirez, P.~Sprechmann, and G.~Sapiro, ``Classification and clustering via
  dictionary learning with structured incoherence and shared features,'' in
  \emph{IEEE Comput. Vision and Pattern Recognition}, 2010, pp. 3501--3508.

\bibitem{bertsekas1999nonlinear}
D.~P. Bertsekas, \emph{Nonlinear Programming}.\hskip 1em plus 0.5em minus
  0.4em\relax Athena Scientific, 1999.

\bibitem{mairal2010online}
J.~Mairal, F.~Bach, J.~Ponce, and G.~Sapiro, ``Online learning for matrix
  factorization and sparse coding,'' \emph{J. of Mach. Learning Research},
  vol.~11, pp. 19--60, 2010.

\bibitem{figueiredo2003algorithm}
M.~A. Figueiredo and R.~D. Nowak, ``An {EM} algorithm for wavelet-based image
  restoration,'' \emph{IEEE Trans. Image Process.}, vol.~12, no.~8, pp.
  906--916, 2003.

\bibitem{daubechies2003iterative}
I.~Daubechies, M.~Defrise, and C.~De~Mol, ``An iterative thresholding algorithm
  for linear inverse problems with a sparsity constraint,'' \emph{Commun. on
  Pure and Appl. Math.}, vol.~57, pp. 1413--1457, 2004.

\bibitem{martinez1998ar}
A.~M. Mart{\'\i}nez, ``The {AR} face database,'' \emph{CVC Tech. Rep.},
  vol.~24, 1998.

\bibitem{martinez2001pca}
A.~M. Mart{\'\i}nez and A.~C. Kak, ``{PCA} versus {LDA},'' \emph{IEEE Trans.
  Pattern Anal. Mach. Intell.}, vol.~23, no.~2, pp. 228--233, 2001.

\bibitem{lecun1989backpropagation}
Y.~LeCun, B.~Boser, J.~S. Denker, D.~Henderson, R.~E. Howard, W.~Hubbard, and
  L.~D. Jackel, ``Backpropagation applied to handwritten zip code
  recognition,'' \emph{Neural computation}, vol.~1, no.~4, pp. 541--551, 1989.

\bibitem{gader1996automatic}
P.~D. Gader and M.~A. Khabou, ``Automatic feature generation for handwritten
  digit recognition,'' \emph{IEEE Trans. Pattern Anal. Mach. Intell.}, vol.~18,
  no.~12, pp. 1256--1261, 1996.

\bibitem{zhou2003ensembles}
Z.-H. Zhou and M.-L. Zhang, ``Ensembles of multi-instance learners,'' in
  \emph{European Conf. on Mach. Learning}.\hskip 1em plus 0.5em minus
  0.4em\relax Springer, 2003, pp. 492--502.

\bibitem{kraut:2001}
S.~Kraut, L.~Scharf, and L.~McWhorter, ``Adaptive subspace detectors,''
  \emph{IEEE Trans. Signal Process.}, vol.~49, no.~1, pp. 1--16, 2001.

\bibitem{facchinei2007finite}
F.~Facchinei and J.-S. Pang, \emph{Finite-dimensional variational inequalities
  and complementarity problems}.\hskip 1em plus 0.5em minus 0.4em\relax
  Springer Science \& Business Media, 2007.

\end{thebibliography}
}
\clearpage

\end{document}